\documentclass{article}
\usepackage{spconf,amsmath,graphicx}
\usepackage{multirow} 
\usepackage{amssymb} 

\usepackage[bookmarks=false,colorlinks]{hyperref}

\usepackage{xcolor}

\usepackage{enumitem}

\usepackage{wrapfig}

\usepackage{adjustbox} 

\title{IMPROVING LIMITED Supervised Foot Ulcer Segmentation Using Cross-Domain Augmentation Strategies}

\name{Shang-Jui Kuo$^{\dagger,\star}$ \qquad Po-Han Huang$^{\dagger,\star}$ \qquad Chia-Ching Lin$^{\dagger}$  \qquad Jeng-Lin Li$^{\dagger}$  \qquad Ming-Ching Chang$^{\ddagger}$
\thanks{$^\star$: The first two authors equally contribute to this work.}}
\address{$^{\dagger}$ Inventec Corporation\\
$^{\ddagger}$ University at Albany, State University of New York\\
\texttt{\footnotesize \{kuo.raysj, huang.po-han, lin.jacob, li.johncl\}@inventec.com}, \texttt{\footnotesize mchang2@albany.edu}
}
\begin{document}
%
\maketitle

\begin{abstract}
Diabetic foot ulcers pose health risks, including higher morbidity, mortality, and amputation rates. Monitoring wound areas is crucial for proper care, but manual segmentation is subjective due to complex wound features and background variation. Expert annotations are costly and time-intensive, thus hampering large dataset creation. Existing segmentation models relying on extensive annotations are impractical in real-world scenarios with limited annotated data. In this paper, we propose a cross-domain augmentation method named {\em TransMix} that combines {\em Augmented Global Pre-training} ({\em AGP}) and {\em Localized CutMix Fine-tuning} ({\em LCF}) to enrich wound segmentation data for model learning. {\em TransMix} can effectively improve the foot ulcer segmentation model training by leveraging other dermatology datasets not on ulcer skins or wounds. {\em AGP} effectively increases the overall image variability, while {\em LCF} increases the diversity of wound regions. Experimental results show that {\em TransMix} increases the variability of wound regions and substantially improves the Dice score for models trained with only 40 annotated images under various proportions. 
\end{abstract}



\section{Introduction}
\label{sec:intro}
\vspace{-3mm}

Foot ulcers, a common complication associated with diabetes, represent a significant public health concern due to their substantial impact on morbidity, mortality, and the elevated risk of lower limb amputations~\cite{DFU_2021}. Precise delineation of wound areas plays a pivotal role in comprehensive wound management and ongoing healing assessment.
Manual skin wound segmentation is a traditionally employed approach that demands a high level of expertise and experience~\cite{wound_seg2017}. This requirement arises from the multifaceted etiologies and diverse visual characteristics of wounds. However, the inherent subjectivity and time-intensive nature of manual segmentation procedures pose considerable challenges within clinical practice~\cite{fugsegnet_challenge_1,ramachandram2022fully}. Consequently, there exists a pressing need to develop an automatic segmentation model for efficient and accurate ulcer wound analysis.

\begin{figure*}[t]
\centerline{
  \includegraphics[width=0.95\textwidth]{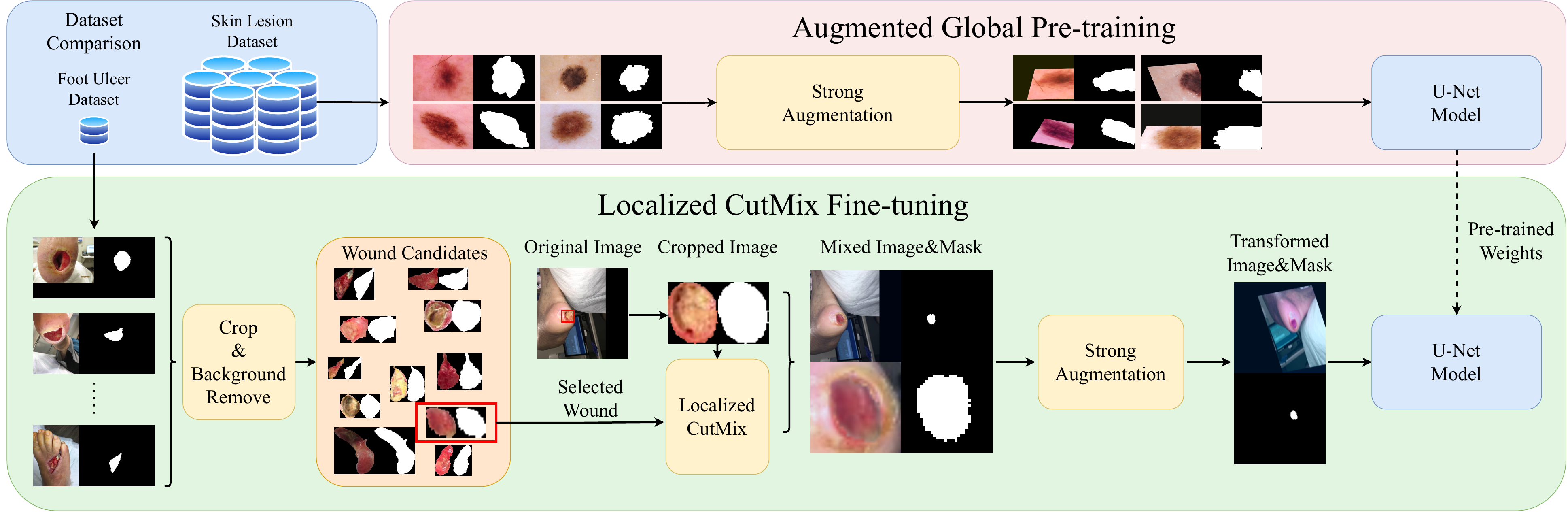}
  \vspace{-3mm}
}
\caption{
{\bf Overview of the proposed {\em TransMix} framework.} 
{\em Augmented Global Pre-training} ({\em AGP}) enables initial model learning from other larger skin datasets. 
{\em Localized CutMix Fine-tuning} ({\em LCF}) expands the diversity of wound patches via CutMix synthesis. 
Our approach can effectively overcome the issue of data scarcity during model training.}
\label{fig_framework}
\vspace{-4mm}
\end{figure*}

Research on automatic wound segmentation has advanced greatly with Deep Neural Networks (DNN), notably the convolutional neural network (CNN) and Transformer architectures. Visual object detection and segmentation networks such as the Mask R-CNN~\cite{mask_rcnn} have been applied for automatic wound segmentation~\cite{automatic_segmentation}. The lightweight model in~\cite{lightweight} utilizes residual attention to capture local patch details. The recent FUSegNet~\cite{fugsegnet_challenge_1} refines the U-Net~\cite{unet} architecture by integrating spatial and channel Squeeze-and-Excitation modules during the decoding phase. In addition, ensemble techniques are broadly employed to overcome the limited discriminative ability of the architecture designs~\cite{fugsegnet_challenge_1,fuseg_challenge_2, ensemble}. 
Despite the advancement, these model designs and ensemble strategies focus on fine-tuning network architectures, without directly manipulating the raw image data. These approaches still face limitations when it comes to effectively learning and segmenting diverse types of wounds. Furthermore, over-parameterized models may struggle to accurately capture subtle wound patterns, especially when there is a scarcity of available labeled training data~\cite{limited_data_2017}.

Previous studies on foot ulcer segmentation have typically neglected real-world scenarios, characterized by a scarcity of labeled training data due to the time-intensive nature of the data collection and annotation processes. In addressing the challenge of data scarcity, cross-domain learning approaches are developed in medical image analysis. Most research focuses on learning from different sources of data that belong to the same set of labeled classes. The domain gap arising from differences in imaging sites for MRI data can be eliminated this way~\cite{bateson2022source}. Other transfer learning approaches extend beyond these domain gaps to encompass different label types. 
For instance, the diabetic retinopathy method is expanded in~\cite{zhou2021cct} to recognize multiple ocular diseases, where their glioma identification model can be used to diagnose other brain tumors. 
In automatic wound segmentation, it is impractical to collect a sufficiently large amount of images with annotations that cover the required variability for DNN model training. 
Although the transfer learning across classes can partially work for wound segmentation, the application-dependent domain gap remains the bottleneck for effectiveness. The main difficulty in developing an automatic foot ulcer segmentation model is still the scarcity of suitable wound image data. To this end, developing effective methods that can learn from limited labeled data is imperative.  

In this work, we propose a cross-domain wound segmentation method based on data augmentation via image transformation and local patch mixing, named {\em TransMix}; see Fig.~\ref{fig_framework}. {\em TransMix} consists of two model training stages: {\em Augmented Global Pre-training} ({\em AGP}) and {\em Localized CutMix Fine-tuning} ({\em LCF}). To verify the effectiveness of learning from limited labels, our model is first trained on the large skin lesion dataset HAM10000~\cite{ham10000, ham10000_mask} from Harvard Dataverse.
The model with learned global knowledge regarding skin segmentation is then fine-tuned on a smaller foot ulcer dataset, where the augmented wound regions with diversified wound appearances can be effectively learned. {\em AGP} enables the learning of global knowledge such as camera viewpoint variations. In {\em LCF}, a more local and refined augmentation using CutMix~\cite{cutmix} is adopted to expand the variability of the wound regions. Instead of randomly replacing patches in the image, we carefully specify the wound region for fine-grained augmentation. The proposed {\em AGP} and {\em LCF} pipeline work jointly to narrow down domain gaps across datasets, enabling our method to robustly transfer the learned knowledge in the scenario of limited annotation data.  
We evaluate our proposed framework on the FUSeg dataset~\cite{fuseg} and achieve an 85.26\% Dice score using only 40 labeled images, which achieved a 10.43\% improvement compared to the baseline method.

\vspace{-3mm}
\section{Method}
\label{sec:method}
\vspace{-2.5mm}

\begin{table*}[t]
\caption{Evaluation of {\em Augmented Global Pre-training} ({\em AGP}) and {\em Localized CutMix Fine-tuning} ({\em LCF}) on the FUSeg validation dataset in terms of precision, recall, and Dice scores in \%.
}
\label{table:main_results}
\centerline{
\setlength{\tabcolsep}{2mm}
\begin{tabular}{c|c|l|ccc|ccc}
\hline
\multirow{2}{*}{\begin{tabular}[c]{@{}c@{}}Experiments\\ Setting \#\end{tabular}} & \multirow{2}{*}{Pre-trained} & \multicolumn{1}{c|}{\multirow{2}{*}{Model}} & \multicolumn{3}{c|}{40 Samples} & \multicolumn{3}{c}{81 Samples} \\ \cline{4-9} 
 &  & \multicolumn{1}{c|}{} & \multicolumn{1}{c|}{Precision} & \multicolumn{1}{c|}{Recall} & Dice & \multicolumn{1}{c|}{Precision} & \multicolumn{1}{c|}{Recall} & Dice \\ \hline
A & - & {\em U-Net} & \multicolumn{1}{c|}{67.99} & \multicolumn{1}{c|}{83.18} & 74.83 & \multicolumn{1}{c|}{\textbf{86.98}} & \multicolumn{1}{c|}{73.24} & 79.52 \\ \hline
B & \checkmark & + Pre-training & \multicolumn{1}{c|}{84.06} & \multicolumn{1}{c|}{55.92} & 67.16 & \multicolumn{1}{c|}{82.98} & \multicolumn{1}{c|}{71.16} & 76.61 \\
C & \checkmark & + {\em AGP} & \multicolumn{1}{c|}{84.99} & \multicolumn{1}{c|}{79.60} & 82.21 & \multicolumn{1}{c|}{84.87} & \multicolumn{1}{c|}{86.50} & 85.68 \\ \hline
D & - & + CutMix & \multicolumn{1}{c|}{80.16} & \multicolumn{1}{c|}{75.16} & 77.58 & \multicolumn{1}{c|}{76.68} & \multicolumn{1}{c|}{\textbf{90.53}} & 83.03 \\
E & - & + CutMix + Background Removal & \multicolumn{1}{c|}{75.11} & \multicolumn{1}{c|}{77.85} & 76.46 & \multicolumn{1}{c|}{84.18} & \multicolumn{1}{c|}{87.29} & 85.71 \\
F & - & + CutMix + Location-aware Pasting & \multicolumn{1}{c|}{85.57} & \multicolumn{1}{c|}{72.38} & 78.43 & \multicolumn{1}{c|}{85.98} & \multicolumn{1}{c|}{86.26} & 86.12 \\
G & - & + {\em LCF} & \multicolumn{1}{c|}{\textbf{86.75}} & \multicolumn{1}{c|}{82.39} & 84.51 & \multicolumn{1}{c|}{84.79} & \multicolumn{1}{c|}{87.53} & 86.14 \\ \hline
H & \checkmark & + {\em TransMix} & \multicolumn{1}{c|}{82.50} & \multicolumn{1}{c|}{\textbf{88.22}} & \textbf{85.26} & \multicolumn{1}{c|}{84.83} & \multicolumn{1}{c|}{88.51} & \textbf{86.65} \\ \hline
\end{tabular}
}
\vspace{-4mm}
\end{table*}

We focus on the wound segmentation task with limited training data. We consider transfer learning from a {\em source} domain with a larger dataset to a {\em target} domain with limited data samples. Specifically, for model pre-training in the source domain, let the source skin dataset $D_S$ consist of image-mask pairs $\{\textbf{x}^S_i,\textbf{m}^S_i\}_{i=1}^M$, where the size $M$ of the source dataset $D_S$ is large enough. Let the target foot ulcer wound dataset $D_T$ consist of image-mask pairs $\{\textbf{x}_i,\textbf{m}_i\}_{i=1}^N$, where $\textbf{m}_i$ denotes the $\mathbb{R}^{W \times H}$ segmentation ground truth of the corresponding image $\mathbf{x}_i$ with the same dimension. We assume the size $N$ of the target dataset to be small ({\em e.g.}, $N\in \{40, 81\}$). The goal is to derive a model that can accurately segment the wound region from the test images in the domain of $D_T$.

\noindent {\bf Dataset.}
We use a real-world foot ulcer wound segmentation benchmark, FUSeg~\cite{fuseg} as the dataset for the target domain. FUSeg contains 810 foot ulcer images along with their ground-truth masks as a training set and 200 images as a validation set. To simulate the data scarcity scenarios for evaluation purposes, we limit the access to only 40 and 81 randomly selected training samples (which corresponds to 5\% and 10\% of the original training set, respectively). 
For the source domain, we utilize the large-scale HAM10000 dataset~\cite{{ham10000}} Harvard Dataverse, as the source dataset. HAM10000 consists of $10,015$ skin lesion images along with ground-truth masks. 

\vspace{-2mm}
\subsection{Cross-Domain Transfer Strategies}
\vspace{-1mm}

Our cross-domain wound segmentation framework in Fig.~\ref{fig_framework} consists of two training stages: {\em Augmented Global Pre-training} ({\em AGP}) and {\em Localized CutMix Fine-tuning} ({\em LCF}). The focus of {\em AGP} is to extract global knowledge of skin lesion segmentation from the larger source-domain data. {\em LCF} then perform location-aware CutMix
with background removal
to enhance wound variability to fine-tune the wound segmentation model on the limited available target-domain samples. Both {\em AGP} and {\em LCF} perform image data augmentations, from coarse to fine, with different purposes: {\em AGP} focuses on global viewpoint variations, while {\em LCF} attends to local wound patch variabilities.

\vspace{-2mm}
\subsection{Augmented Global Pre-training}
\label{sssec:global_transfer}
\vspace{-1mm}


To mitigating the large discrepancy between the source and target domains, effective data augmentation is necessary for knowledge transfer during model training.
As shown in the snapshots in Fig.~\ref{fig_framework}, HAM10000 contains higher-quality images taken with consistent camera view angle and distance, illumination conditions and backgrounds. 
The skin lesion regions are mostly circular in shape, occupying a large portion at the center. In contrast, many images in the FUSeg dataset are taken with various conditions and qualities, with arbitrary wound locations, shapes and sizes.
To this end, we perform intense spatial-level transformations including global perspective transformations and optical distortion. We also perform strong pixel-level transformations including RGB \& HSV shifts, brightness \& contrast, to simulate illumination changes. A U-Net~\cite{unet} model is pre-trained to learn the segmentation of skin regions, which will be transferred to the foot ulcer domain for wound segmentation.

\vspace{-2mm}
\subsection{Localized CutMix Fine-tuning}
\label{sssec:localized_augmentation}
\vspace{-1mm}

With the pre-trained model using the large-scale HAM10000 dataset, we perform fine-tuning to the FUSeg dataset for wound segmentation. 
In a situation with very limited training data, the model can hardly capture diverse types of wounds. Therefore, our proposed {\em TransMix} approach includes a CutMix strategy to diversify the visual appearances of the wounds during the fine-tuning steps.

We first obtain the bounding boxes' position of the wounds in all the training images using the corresponding annotated masks and then crop the bounding boxes along with their masks as the wound candidates. The background of each wound candidate is removed based on the corresponding mask to synthesize a more realistic wound appearance without an obvious artificially generated cropping boundary.

In each batch of data, we specify the position of the wound in target images and randomly paste other wound candidates on the specified region. As the wound size can be dramatically different across images, we deliberately select the wound candidates with similar wound sizes to the target wound region and resize the wound candidates to match the target wound size.
When we paste a wound candidate to a specified region, part of the wounds in the target image remains, and other parts of the pasted wound candidate might exceed the specified boundary due to the different shapes. We regard the union regions of the remained wounds and the pasted wound candidates to generate a new segmentation mask for the newly synthesized image.
Through multiple iterations in the fine-tuning phase, the same wound region is augmented with multiple wound candidates which enhances the robustness of segmentation model learning against various wound appearances. Aside from the localized wound CutMix, we also perform the strong whole-image transformation mentioned in~$\S$\ref{sssec:global_transfer} to augment the training data perspectives.

\vspace{-1mm}
\section{Experimental Results}
\label{sec:experiment}
\vspace{-2mm}

We set up two experiments to evaluate our proposed framework in terms of the comparison to other segmentation approaches and the analysis of different numbers of accessible training data.
The results are reported in~$\S$\ref{exp1} and~$\S$\ref{exp2}, respectively. We include the compared models and implementation details in the following sections.



\noindent {\bf Models for comparison.}
The settings of the compared models for~$\S$\ref{exp1} are described below and correspond to the results in Table~\ref{table:main_results}. 
We adopt \emph{U-Net} (Setting A) as a strong baseline model which has been widely used for automated wound segmentation~\cite{detect-and-segment}. Notably, FUSegNet~\cite{fugsegnet_challenge_1} using an EfficientNet-B7~\cite{efficientnet} backbone with ensemble techniques, achieved a state-of-the-art Dice score (92.70\%) on the FUSeg dataset. We kept the architecture but avoided such a cumbersome backbone model for the data-scarce scenario while still achieving a comparable Dice score (91.08\%) on the full set evaluation with the ResNet-50 backbone. 

\emph{Pre-training} (Setting B) is performed using the HAM10000 dataset with naive fine-tuning. This baseline is a common practice in the medical field in dealing with limited training data~\cite{peng2021medical} although most past studies are limited to using segmentation datasets that have the same set of labeled classes. \emph{AGP} (Setting C) utilizes our proposed pre-training approach (~$\S$\ref{sssec:global_transfer}) using naive fine-tuning.

We also compare model training approaches without pre-training as follows. CutMix (Setting D) is applied based on the original paper~\cite{cutmix}. We further remove the background of the wound candidates and specify the wound location in Setting E and Setting F. {\em LCF} (Setting G) is our proposed approach in~$\S$\ref{sssec:localized_augmentation} without pre-training.

Ultimately, our proposed \emph{TransMix} (Setting H) combines both {\em AGP} and {\em LCF}.

\noindent {\bf Implementation Details.}
We use the dataset described at ~$\S$\ref{sec:method} and U-Net with ResNet-50 encoder as our backbone.
We used Adam optimizer with 5e-4 learning rate, batch size of 4, and training with 300 epochs, the learning rate is decayed by a factor of 10 every 100 epochs. We perform data augmentation at two levels using the widely-used toolbox of  Albumentations~\cite{albumentation}. At the spatial level, we use several augmentations including grid distortion, optical distortion, perspective transform, and affine transform to align the whole image angle across the dataset.
At the pixel level, we applied some transformations such as RGB shift, brightness contrast, and HUE saturation value transformations to align illumination conditions.

\begin{figure}[t]
\centerline{
  \includegraphics[width=1.2\linewidth]{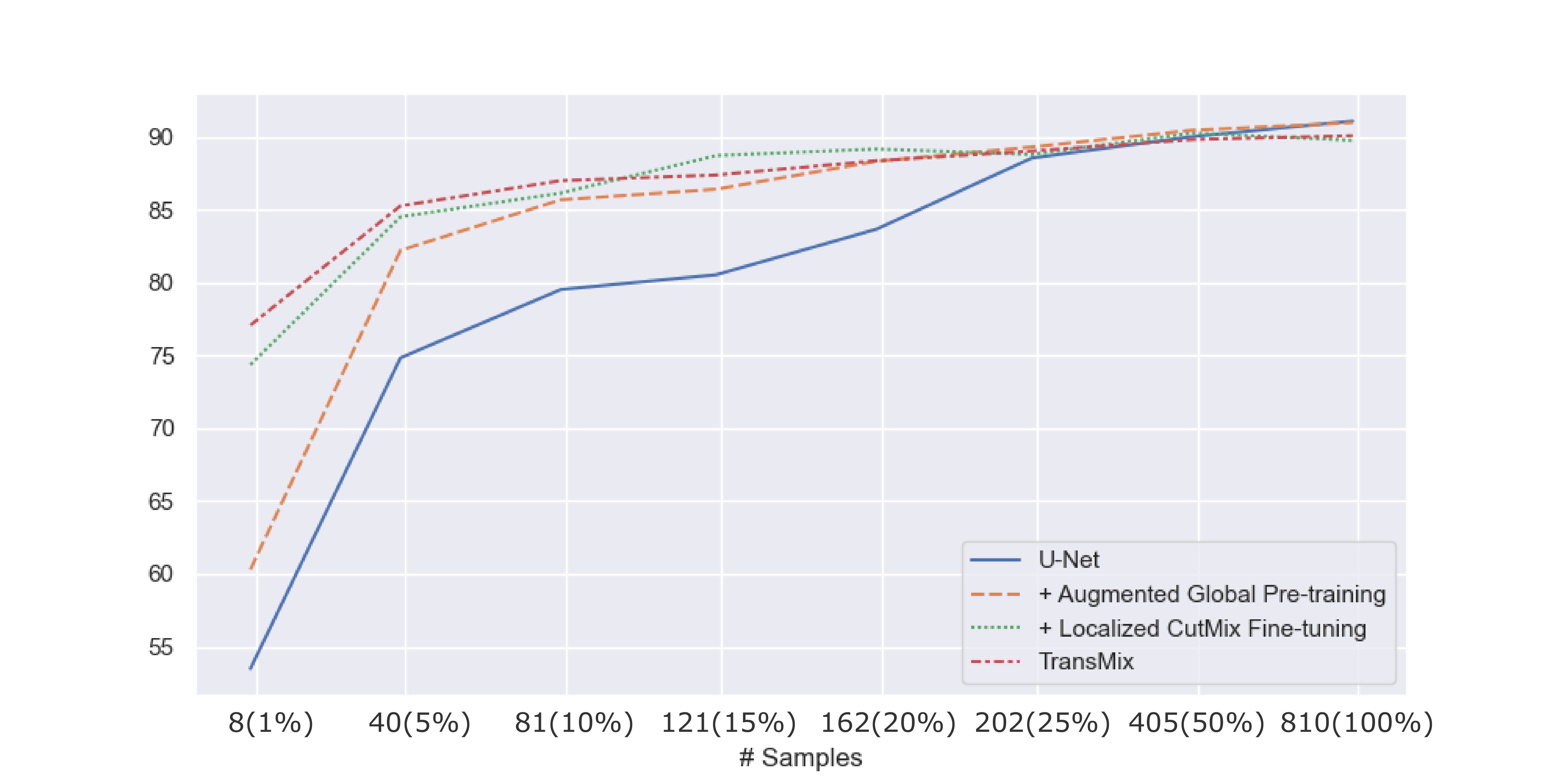}
  \vspace{-2mm}
}
\caption{We evaluate the model fine-tuned using different numbers of FUSeg training samples, and report the Dice scores (\%) on the FUSeg validation set.}
\label{fig:data_amount}
\vspace{-2mm}
\end{figure}

\vspace{-1mm}
\subsection{Results Under Limited Supervision}
\label{exp1}
\vspace{0mm}

We compare our proposed {\em TransMix} method against other segmentation approaches under limited supervision.
Table~\ref{table:main_results} lists the performance comparisons on the FUSeg validation set using only 40 or 81 training samples. As can be seen, our framework ({\em TransMix} in Setting H) consistently demonstrates significantly superior performance compared to baseline methods. Notably,
our framework achieves Dice scores of 85.26\% and 86.65\%, respectively, showing 10.43\% and 7.13\% improvements compared to the baseline method (\textit{U-Net} in Setting A).


\noindent {\bf Effects of the Augmented Global Pre-training.} 
As mentioned in~$\S$\ref{sssec:global_transfer}, it is essential to eliminate the domain gap between the source and target domains for effective knowledge transferring.
As can be seen from Setting B in Table~\ref{table:main_results}, simple pre-training on HAM10000 for transfer learning could result in significant performance degradation as compared to the baseline (Setting A).
In contrast, the model pre-trained by the proposed {\em Augmented Global Pre-training} strategy (Setting C) achieves Dice scores of 82.21\% and 85.68\% after fine-tuning on 40 and 81 FUSeg training samples, respectively, showing 7.38\% and 6.16\% improvements compared to the baseline.
This demonstrates the necessity of our strong augmentation strategy in eliminating the domain gap to better benefit from the pre-training process, especially in data-scarce scenarios.


\noindent {\bf Effects of the Localized CutMix Fine-tuning.}
As mentioned in~$\S$\ref{sssec:localized_augmentation}, it is crucial to synthesize diverse wound samples with reasonable locations and realistic appearances for useful data augmentation. We conduct an ablation study to demonstrate the effectiveness of the proposed {\em LCF} in fulfilling this goal.
Settings D-G in Table~\ref{table:main_results} list the performance comparisons under different CutMix strategies during fine-tuning.
We employ the original CutMix~\cite{cutmix} technique (Setting D), and further remove the background of wound candidates to be pasted (Setting E). Both of these settings show improvements compared to the baseline (Setting A). However, since the position for pasting is randomly selected, these settings may result in unrealistic synthesized images.
Furthermore, we paste wound candidates based on the wound locations in the target image without performing background removal (Setting F) to verify the effectiveness of location-aware pasting. This approach resulted in an improvement compared to Setting D, highlighting the importance of a reasonable pasting position.
Finally, we simultaneously employ background removal and location-aware pasting (Setting G), which achieves further improvements of 6.93\% and 3.11\% in Dice scores, respectively.

\vspace{-1mm}
\subsection{Analysis on the Amount Training Data}
\label{exp2}
\vspace{1mm}

In Fig.~\ref{fig:data_amount}, we examine the change of Dice score with varying data volumes, using 40 (5\%), 81 (10\%), 202 (25\%), 410 (50\%), and 810 (100\%) images to simulate the data accessibility in the real world. Our proposed {\em TransMix} brings about large benefits when we only obtain less than 200 wound samples. The trend shown in Fig.~\ref{fig:data_amount} suggests the benefits of {\em TransMix}, while the margin becomes minor if the target dataset is large enough. With very limited labeled data (e.g., $N<40$), the model performance could highly depend on the quality of the sampled data. 

\vspace{1mm}
\section{Conclusion}
\label{sec:conclusion}
\vspace{-3mm}

Diabetic foot ulcers patients require continual wound monitoring to manage their recovery conditions. However, current automatic segmentation approaches assume the accessibility of large-scale labeled data, which is impractical in clinical practice. To our best knowledge, this work is the first study addressing the limited supervised wound segmentation problem. The novel cross-domain framework enables the use of other datasets with skin images for global knowledge transfer and augments wound variability in a localized manner. 

{\bf Future work:} we will apply few-shot and zero-shot learning to enhance model usability in the real world.

\clearpage

\bibliographystyle{IEEEbib}
\bibliography{refs_simple}

\end{document}